# W-Net: Reinforced U-Net for Density Map Estimation

Kinal Mehta[*]
C-DAC, Pune
kinalm@cdac.in

Varun Kannadi Valloli[*]
C-DAC, Pune
varunkv@cdac.in

## Abstract

*Crowd management is of paramount importance when it comes to preventing stampedes and saving lives, especially in countries like China and India where the combined population is a third of the global population. Millions of people convene annually all around the nation to celebrate a myriad of events and crowd count estimation is the linchpin of the crowd management system that could prevent stampedes and save lives. We present a network for crowd counting which reports state of the art results on crowd counting benchmarks. Our contributions are, first, a U-Net inspired model which affords us to report state of the art results. Second, we propose an independent decoding Reinforcement branch which helps the network converge much earlier and also enables the network to estimate density maps with high Structural Similarity Index (SSIM). Third, we discuss the drawbacks of the contemporary architectures and empirically show that even though our architecture achieves state of the art results, the merit may be due to the encoder-decoder pipeline instead. Finally, we report the error analysis which shows that the contemporary line of work is at saturation and leaves certain prominent problems unsolved.*

## 1. Introduction

Crowd monitoring has always been integral to running a safe event. Historical analysis of the crowd monitoring data could be used to understand the crowd and help work out how many attendees to expect and plan events accordingly. Crowd counting can help with both monitoring and storing relevant historical data. In this paper, we not only present our attempt at solving the crowd counting problem, but also discuss our observations on the flow of contemporary line of work. Crowd counting is not a trivial problem and brings along various obstacles, like, occlusion, background noise and variations in illumination, distribution of people, scale and perspective. Solutions have come a long way since Lempitsky et al. [1] in tackling some of these issues. MCNN [2] attempts to tackle the problem with a multi-branch architecture to handle scale variation and SANet [3] builds on the inception architecture to handle scale and build high resolution density maps to estimate accurate crowd count. Even though these architectures have produced incrementally better results, they have done so by focusing on one major obstacle in the crowd counting problem, i.e., scale-variance. On the other hand, some have tried to tackle the problem by data manipulation, generating density maps with adaptive gaussian kernels based on head detections [4] and improving crowd counting using inverse k-Nearest Neighbor maps [5]. We definitely have come a long way since Lempitsky et al. [1], but these independent efforts solve only the specific issues they are tackling. Considering this, we attempt to solve the crowd counting problem by focusing on scale-invariance and Reinforcement for background noise reduction, retaining structural similarity and improving convergence.

SANet [3] reported state of the art results in 2018 and the output was primarily credited to the inception encoder module and high-resolution density map generated. The odd thing is that high resolution density map is not a documented necessity for accurate crowd counting. CSRNet [6] generates a density map that is at an eighth resolution compared to the input and its results on the crowd counting benchmark are only marginally worse than that of SANet. The addition of extra decoding layers in SANet alone makes it computationally heavy, ignoring the inception encoding module which is heavier. The upsampling in SANet is taken care of by the Transpose layers which adds checkerboard artifacts into the mix making it a soup sandwich. We have attended to this problem using Nearest neighbor interpolation as an alternative, as suggested in [7].

Scale-Aware Attention Network [4] presented an interesting multi-branch architecture with a soft attention mechanism that learnt a set of gating masks. Inspired by this mechanism, we implemented a "Attention" decoding branch trained as a classifier in our architecture to reinforce the final estimated density map. This branch, named "Reinforcement branch", is NOT an attention mechanism, rather, it is just a mechanism for the network to converge faster. Fig 3. shows the architecture of the Reinforcement branch. The Reinforcement branch and the Density map estimation (DME) branch are almost identical, except for the fact that the Reinforcement branch has an extra Conv1_1 layer followed by a Sigmoid layer. This extra convolution layer does not have ReLU or Batch Normalization.

The proposed model is inspired by U-Net [8], we build a network on the same encoder-decoder pipeline with the additional Reinforcement branch. The encoder block of the U-Net is replaced by VGG16bn [9]. We empirically found that VGG16bn was the best backbone to use among VGG16bn, RESNET50[10] and an Inception [11] inspired feature extractor. In our experiments, we found that training the network without the Reinforcement branch drastically slowed down the convergence. This was the main reason the Reinforcement branch was introduced as a binary classifier. The reinforcement network, when independently trained, converged quicker. but was useless in regards to crowd counting, whereas the Density Map Estimation (DME) branch was harder to train, but counted well. We combined the two with a joint loss, i.e., Binary cross entropy (BCE) Loss and Mean squared error (MSE) Loss and trained the network in an end to end fashion.

---
[*]Equal Contribution. Name ordering determined by a game of Jan-Ken-Pon.

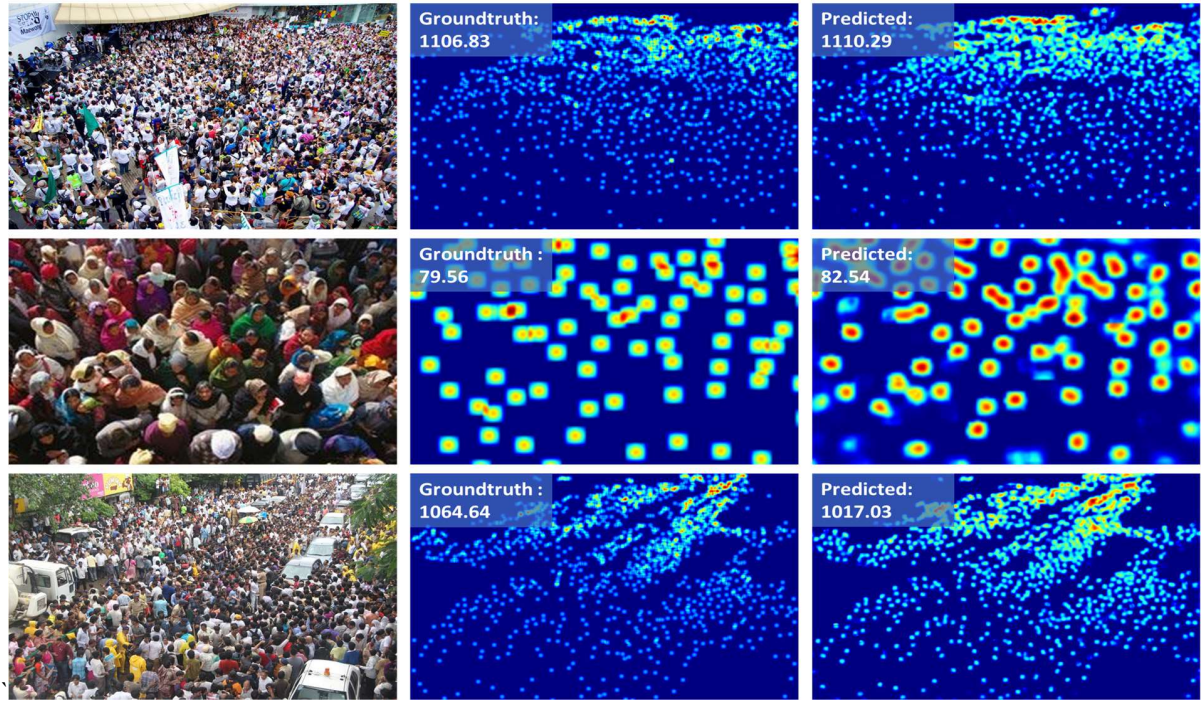

Fig. 1: Model output with groundtruth density maps

To summarize, we make the following contributions: (i) We propose a modified network architecture that generates accurate density maps at half resolution compared to the input. (ii) We propose a Reinforcement branch for the network to converge quicker. (iii) Reporting results on crowd comparison benchmarks and comparing against competing approaches. (iv) Extensive ablation studies are conducted to show effectiveness of DME and Reinforcement branches working together, efficacy of VGG16bn as the backbone over ResNet50 and Inception modules, and replacing transpose layer with Nearest neighbor upsampling. (v) We finally discuss our error analysis and show the drawbacks of the current methods.

## 2. Related Work

A large variety of methods have been proposed to tackle the crowd counting problem. We can largely be bifurcated as conventional approaches and CNN based approaches.

### 2.1. Conventional Approaches

Early research on crowd counting primarily was focused on detection based counting [12]. Occlusion is the biggest obstacle in such solutions and the best you can do to minimize error is to detect smaller parts of the pedestrian like a face or head rather than an entire body. Even while detecting heads, in a densely crowded scenario, the scale variation makes it quite hard for the detection model to detect entities throughout the image. Another approach was to extract features from an image and map it to the count of people. These methods were almost entirely dependent on the feature extraction process and were limited to the adaptability of the feature extraction algorithm to test images. Similar regression based approaches built on this work and added features extracted from Fourier transforms [14], detections [12] and SIFT [13] to regress the count in an image. The main drawback of this line of work was that the global count was predicted but it ignored spatial and sematic information in the images. The importance of using both spatial and semantic information to regress the count is discussed in this TED encoder-decoder approach [15].

### 2.2. CNN-based approaches

CNN-based approaches have shown great potential at solving crowd counting credited primarily to their representation learning ability and robustness. Wang et al. [16] proposed a solution where a modified AlexNet [17] was used directly for predicting count. Contemporary CNN based approaches mainly revolve around density map estimation and regressing the crowd count from the density map. Zhang et al. [18] proposed a CNN trained to estimate the density map and then regress the crowd count. Following this pattern, Walach and Wolf [19] attempted to better the system by using layered boosting and selective sampling. Shang et at.[20] proposed a network to predict local and global count from images. Zhang et al. [2] built a multi branch network dubbed Multi-column CNN (MCNN) to handle scale variations. Even though this work is commendable, it is limited to the number of branches in the network. Boominathan et al. [22] proposed a density map estimator which was a combination of shallow and deep networks. Sam et al. [23], inspired by MCNN, proposed Switch-CNN, an approach where instead of having one network handle input with scale variation, a classifier first classifies an image patch to select an appropriate regressor for the input scale. Li et al. [6] proposed a encoder decoder network, CSRNet, where the backbone was a pretrained VGG16[9] network and decoder was built with dilated convolution later. Even though the estimated density map was of an eighth resolution of the input, CSRNet was able to report state of the art results. A similar encoder decoder model, Xinkun et al. [3] proposed SANet which used inception modules in the encoder to handle scale variation and extract features and Transposed convolution layers in the decoder to upscale the feature map extracted. This method was able to build high resolution density map and beat CSRNet, albeit a marginal improvement.

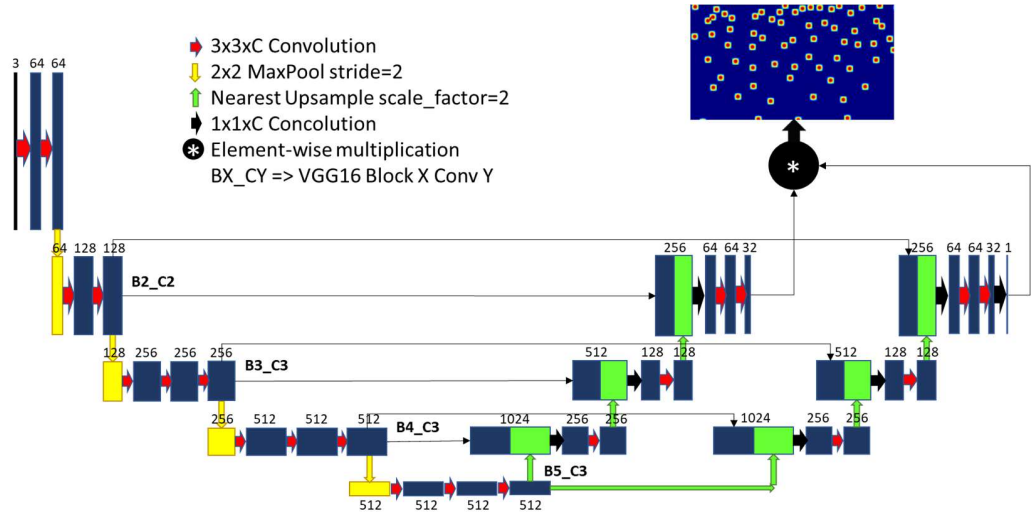

Fig. 2: W-Net Model Architecture

We observed that the most of the state-of-the-art contemporary approaches used the encoder decoder pipeline. They primarily focused on a backbone feature extractor which was scale invariant. As discussed in section 1, we empirically found that VGG16bn worked the best for us as the feature extractor. As for the decoder, our proposed system outputs a density map that is at half the resolution of the input image. The resolution of the density map may or may not directly affect the results as a high-resolution map, like the one generated by SANet, only produces marginally better results than the low resolution one produced by CSRNet. The extra layer may introduce unnecessary complications and delay convergence of the network. We also attempt to reinforce the network with a Reinforcement decoder branch for faster convergence and also to keep a check on local pattern consistency.

3. W-Net

This section presents the details on W-Net. We will introduce network architecture first and then discuss the constituent branches in detail.

3.1. Architecture

As shown in Fig 2, our architecture is inspired from U-Net [8]. U-Net was designed for biomedical image segmentation. The architecture consists of a "contracting path" to capture context and a symmetric "expanding path" that enables precise localization. This encoder-decoder structure was what we built our network on. The W-Net has 3 branches, first the encoder branch, followed by a split in the network which flows parallel to the Density Map Estimation (DME) branch and the Reinforcement branch. The Encoder branch extracts the multi-scale feature maps and the DME branch outputs the density map. The Reinforcement branch is used as to construct an auxiliary input which helps the network converge faster and keep local pattern consistency.

**Encoder Branch:** The encoder branch is the feature extractor in the encoder-decoder pipeline. We chose VGG16bn as our backbone after experiments on Shanghai Part B dataset with VGG16bn, ResNet50 and Inception based feature extractor as the encoder. The results of this experiment is discussed in section 6.2. Our results were corroborated by the choice of front-end for CSRNet which followed ideas proposed in [22,23,30]. We only use the feature extractor from VGG16bn and remove the classification section (fully connected layers). We carved out the first 13 layers from the pretrained VGG16bn from torchvision as our backbone. We use feature maps from B2_C2, B3_C3, B4_C3 and B5_C3 as shown in Fig. 2 as inputs to our decoder branch. Following U-Net, these inputs at different abstract levels help represent multi-scale features.

**DME and Reinforcement branch:** The Decoder branches, i.e., DME and Reinforcement branch have a common structure which is described in Fig. 3. In this subsection, we will discuss the common structure first and then follow it up with the explaining the differences. First, the output of B5_C3 is upscaled using Nearest neighbor interpolation and then concatenated with the output of B4_C3. This concatenated input is given to Decoder Block 1 described in Fig 3. Decoder block 1 contains a conv1×1×256 and conv3×3×256. The output of Decoder Block 1 is upscaled and concatenated with the output of B3_C3. The same upscaling protocol is repeated before being fed to Decoder Block 2 which has a similar structure but is different only in terms of channel size which is described in Fig. 3. Finally, after another upscaling and concatenation with B2_C2, Decoder Block 3 outputs a feature map which is the final 32 channel output from the common structure for DME and Reinforcement branch. Now for the differences, the Reinforcement branch feeds the final output from Decoder Block 3 to a conv 1×1×1 followed by Sigmoid activation. This generates the Reinforcement map which is element-wise multiplied to the final output from Decoder Block 3 of the DME branch. This is fed to the last conv1×1×1 layer with ReLU activation which generates the final density map.

### 3.2. Loss Functions

For the loss function, we use a joint loss which consists of Mean Squared Error (MSE) and Binary Cross Entropy (BCE) loss. SANet uses SSIM loss to make sure that local correlation of the density map is not ignored, but in our case, we found that the Reinforcement branch, which is a classifier, converges quickly and can be used to influence the density map estimation branch. This way, we not only help the network converge quicker, but also helps us keep a check on the local correlation of the density maps. Table 5 compares the Structural similarity of the density maps estimated with different methodologies and it clearly shows how structural integrity is maintained in our method.

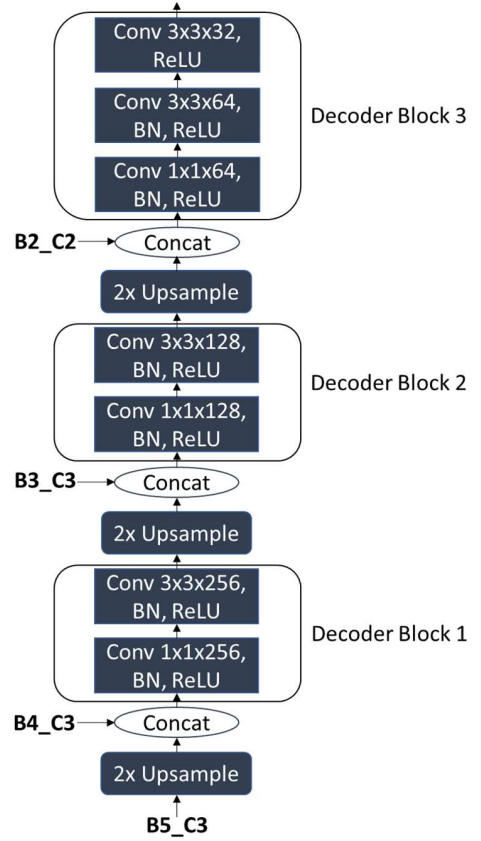

Fig. 3: Decoder Branch Common Architecture

**MSE loss:** We chose MSE loss because of the insights shared by Hang et al. [24] in their study of Loss Functions for Image Restoration. One interesting property they shared was the fact that $L_2$ provides the maximum likelihood estimate in case of independent and identically distributed Gaussian noise. SANet points to the fact that most contemporary work use Euclidean distance which is based on the pixel independence hypothesis and ignores the local correlation of the density maps. The reinforcement branch in our architecture is used to handle this issue and makes MSE loss a viable loss function.

$$MSE(y, t) = \frac{1}{N}\sum_{i=1}^{N}(y_i - t_i)^2 \qquad (1)$$

where y is the predicted value, t is the target and N is the total number of pixels.

**Binary Class entropy:** For the reinforcement branch, we used BCELoss for training this classifier. Binary cross entropy loss is defined as follows:

$$BCE(y,t) = \frac{1}{N}\sum_{i=1}^{N}[t_i \times \log(y_i) \qquad (2)$$
$$+ (1 - t_i) \times \log(1 - y_i)]$$

where y is the predicted value, t is the target class and N is the total number of pixels.

The network is trained in an end to end fashion with a joint loss function L which is defined as:

$$L = (\alpha \times L_{MSE}) + (\beta \times L_{BCE}) \qquad (3)$$

where a and b are used to balance the loss values and we empirically found that α=1000 and β=10 works optimally for training the network.

4. Training and evaluation details

In this section we discuss the details of the training and evaluation procedures.

**Density map generation:** We generate density maps following the method described in [3]. We convert the ground truth to density maps by, convolving a Normalized Gaussian kernel over delta function δ(x-x$_i$) where xi is a targeted object.

$$F(x) = \sum_{i=1}^{N}\delta(x - x_i) \times G_{\sigma_i}(x) \qquad (4)$$

We also use a fixed spread parameter σ of the gaussian kernel to generate the density maps. In our experiments, we used a kernel of window size (μ)=15 and spread parameter (σ)=4.

We attempted geometry-adaptive kernels as described in the work by Zhang et al. [6], where the spread parameter σ was a variable dependent on the average distance of K-Nearest Neighbor. We define σ for each head to be

$$\sigma_i = \beta \overline{d_i} \qquad (5)$$

where β = 0.3 and 'i' is each head annotation. We found that the results for our network was considerably poorer when compared to using a fixed spread parameter. The comparison of the results based on the dataset generation is shown in Table 1.

| Method | MAE |
|---|---|
| Geometry-adaptive kernels | 79.3 |
| **Fixed sigma kernel** | **59.5** |

Table 1: Data generation result comparison

**Reinforcement map generation:** For the reinforcement maps, we used the same method as density generation but used a larger window size and spread parameter. Once the blurred map is generated, we use binary thresholding to create classification map to train the reinforcement branch. In our experiments, we used a threshold (th) = 0.001 and the classification map was computed using Equation (4).

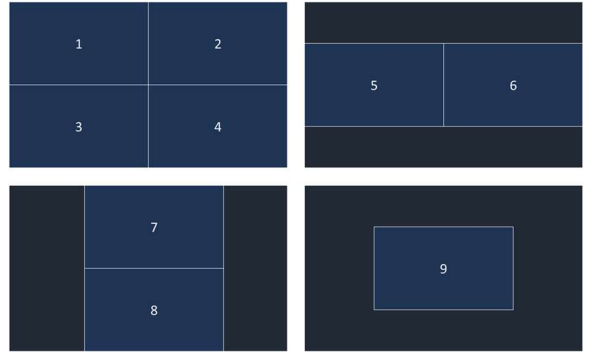

Fig. 4: Nine patches for patch-wise evaluation

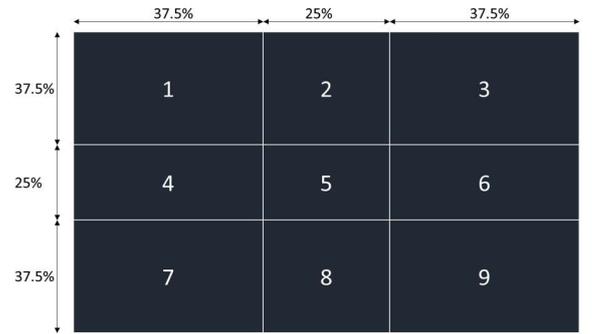

Fig. 5: Percentage contribution from respective patches.

4.1. Training details

We follow the patch based training similar to SANet, the difference is that we used fixed size patches to generate training batches. In our experiments, we used 400X400 image patches cropped at random locations to create batches of 14 for training. The only data augmentation we used was to randomly flip the image horizontally half the time.

The implementation of W-Net is done in PyTorch [25] framework. For the encoder branch in W-Net, we use a pretrained VGG16bn model from torchvision. The decoder branches are randomly initialized by Gaussian distribution with mean zero and standard deviation of 0.01. We use Adam optimizer [26] with a learning rate of 1e-4 and weight decay of 5e-3. Our attempts with Stochastic gradient descent with momentum took much longer to converge and hence was dropped. We use batch normalization layers after

every convolution layer except for the output layers, this is detailed in Fig. 2 and Fig. 3

4.2. Evaluation Details

For the evaluation of our models, we use patch based evaluation as described in [3]. We crop the images into quarters and generate nine overlapping quarters for an image as portrayed in Fig 4. Each quarter has its density map predicted and the final output is made of the merged output maps. Fig. 5 shows which patch contributes how much percentage of the final Density map. The corner image patches contribute the most, i.e., since they have two edges without overlaps, that patch is the sole contributor to those edges. The image patches numbered 2,4,6 and 8 contribute the second most, as they have one edge without overlaps. Finally, image patch 5 contributes the least because all of its edges have overlaps.

As metrics to evaluate our crowd counting models, we use Mean Absolute error (MAE) and Root Mean Squared Error (RMSE). They are defined as follows:

$$MAE = \frac{1}{N}\sum_{i=1}^{N}|C_i - C_i^{GT}| \quad (6)$$

$$RMSE = \sqrt{\frac{1}{N}\sum_{i=1}^{N}(C_i - C_i^{GT})^2} \quad (7)$$

where C is the predicted count while $C^{GT}$ is the ground truth count and N is total number of images.

MAE is used to indicate how accurate the result is and RMSE is used to gauge the robustness of the model.

5. Experiments

In this section, we present our results on three public crowd counting benchmark datasets, namely, ShanghaiTech [2], UCF_CC_50 [27] and UCSD [28].

5.1. ShanghaiTech dataset

The ShanghaiTech dataset [2] consists of 1198 images, with 330,165 annotated people. This dataset is bifurcated in two parts: Part A and Part B. Part A contains 482 images with a train split of 300 images and a test split of 182 images which is provided by the authors. Part B contains 716 images, out of which 400 are for training and 316 images for testing. Shanghai Part A consists of random images crawled from the internet, and Shanghai Part B contains images captured from street view. The ground truth maps and Reinforcement maps along with the augmented data are generated as discussed in section 4. Table 2 shows the comparison of the results of our method with contemporary state of the art works on Shanghai Part A and Shanghai Part B datasets. We report significant improvement over the State-of-the-art method SANet.

|  | Part A | | Part B | |
| --- | --- | --- | --- | --- |
| Method | MAE | MSE | MAE | MSE |
| Zhang et al. [18] | 181.80 | 277.70 | 32.00 | 49.80 |
| MCNN [2] | 110.20 | 173.20 | 26.40 | 41.30 |
| Switch-CNN [23] | 90.40 | 135.00 | 21.60 | 33.40 |
| CP-CNN [29] | 73.60 | 106.40 | 20.10 | 30.10 |
| CSRNet [6] | 68.20 | 115.00 | 10.60 | 16.00 |
| SANet [3] | 67.00 | 104.50 | 8.40 | 13.60 |
| **W-Net(ours)** | **59.50** | **97.30** | **6.90** | **10.30** |

Table 2: Results on ShanghaiTech Dataset

5.2. UCF_CC_50

UCF_CC_50 is a small dataset which consists of 50 annotated images. There is a large variation in the crowd density, from 94 to 4543 people in an image. The limited size of the dataset makes this a challenging problem to solve. We use 5-fold cross-validation to evaluate our model on this dataset as described in [3]. The ground truth maps and Reinforcement maps along with the augmented data are generated as discussed in section 4. Table 3 shows the comparison of the results of our method with contemporary state of the art work on UCF_CC_50 dataset. W-Net beats SANet, which is the state-of-the-art method, by a significant improvement.

| Method | MAE | MSE |
| --- | --- | --- |
| Idrees et al. [27] | 419.5 | 541.6 |
| Zhang et al. [18] | 467 | 498.5 |
| MCNN [2] | 377.6 | 509.1 |
| Switch-CNN [23] | 318.1 | 439.2 |
| CP-CNN [29] | 295.8 | 320.9 |
| CSRNet [6] | 266.1 | 397.5 |
| SANet [3] | 258.4 | 334.9 |
| **W-Net(ours)** | **201.9** | **309.2** |

Table 3: Results on UCF_CC_50 Dataset

5.3. UCSD

The UCSD dataset is a much larger dataset which consists of 2000 frames from a surveillance video camera. The resolution of the frames is quite low and so is the density of the people. The number of people averages around 25 for a frame. UCSD Provides Region of Interest to mask the background and make the counting task simpler. The train test split for this dataset is as described in [28], i.e., frames 601 to 1400 are used for training and the remaining are used for testing. Although the generation of ground truth maps and Reinforcement maps are as described in section 4, we have preprocessed the images to have a resolution of 400X602. The shorter edge is resized to 400 and the longer one to 602 to maintain aspect ratio. This lets us use our training pipeline without any modification for this dataset as we use 400X400 patches to train. Table 4 shows the results and W-net manages to beat SANet even in sparse crowds.

| Method | MAE | MSE |
|---|---|---|
| Zhang et al. [18] | 1.6 | 3.31 |
| MCNN [2] | 1.07 | 1.35 |
| Huang et al. [21] | 1 | 1.4 |
| Switch-CNN [23] | 1.62 | 2.1 |
| CSRNet [6] | 1.16 | 1.47 |
| SANet [3] | 1.02 | 1.29 |
| **W-Net(ours)** | **0.82** | **1.05** |

Table 4: Results on UCSD Dataset

6. Ablation Studies

6.1. Validation for Reinforcement Branch

In this subsection, we perform ablation study to analyze the efficacy of the Reinforcement branch. The experiments are done on Shanghai Part A dataset. We claim that the reinforcemnt branch not only speeds up convergence, it also keeps a check on the structural similarity of the ground truth density maps and the estimated density maps. To measure stuctural similarity, we use Structural Similarity Index (SSIM) as metric to gauge the structural integrity. As shown in Table 5, we can see how the produced density maps have a significantly higher SSIM value than the previously documented methods. As for the speed, the model converged twice as faster with the Reinforcement branch than with just the density estimation branch.

| Method | SSIM |
|---|---|
| MCNN [2] | 0.52 |
| CP-CNN [29] | 0.72 |
| CSRNet [6] | 0.76 |
| **W-Net(ours)** | **0.93** |

Table 5: SSIM Comparison

| Method | MAE |
|---|---|
| **VGG16bn** | **6.998** |
| ResNet50 | 8.5 |
| Inception based | 10.2 |

Table 6: Comparison of encoders

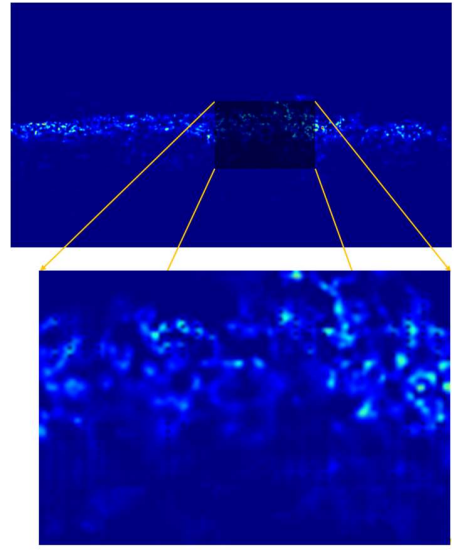

Fig. 6: Checkerboard artifacts due to transpose convolution layers

6.2. Validation for VGG as the encoder

In this subsection, we present our findings on experimenting with VGG16bn, ResNet50 and Inception based architecture as encoders. Each of these architectures handle scale variance in their own ways, and we empirically found that VGG16bn produced the best results for this use case, both computationally and result-wise. Table 6 shows the result of the encoder-decoder pipeline with these encoders respectively on Shanghai Part B dataset.

6.3. Ablation experiments on upsampling layers

In this subsection, we present our observations of the benefits of using nearest neigbor interpolation versus transpose convolution layers for upsampling. SANet partially credits its state-of-the-art results to the high-resolution density maps it generates. SANet uses transpose convolution layers to upsample its encoded feature maps. The problem with transpose convolution layers is that the upscaled images it generates present themselves with checkerboard artifacts. Referring to [7], Nearest neighbor interpolation is an alternative which is not affected by the checker board artifact. In our experiments, our reimplementation of SANet present with checkerboard artifacts and so did W-Net. The models also took much longer to converge and when they did, the performance was much worse. Fig. 6 shows the checkerboard artifacts when trained with transpose convolution layers and Fig. 1 shows that the density map estimated with nearest neighbor interpolation is not affected with checkerboard artifact.

| File Name | Our Model MAE | CSRNet MAE |
|---|---|---|
| IMG_8 | 444 | 533 |
| IMG_165 | 424 | 430 |
| IMG_127 | 380 | 269 |
| IMG_122 | 377 | 313 |
| IMG_90 | 361 | 433 |
| IMG_36 | 285 | 291 |
| IMG_110 | 252 | 202 |
| IMG_63 | 198 | 155 |
| IMG_54 | 161 | 326 |
| IMG_173 | 145 | 196 |

Table 7: Worst Performing Images

## 7. Error Analysis

In this section, we discuss our observations on the results of CSRNet and W-Net on Shanghai Part A. We chose CSRNet to accompany our model in this analysis because out of all the encoder decoder models, CSRNet was the most different, architecture wise and result-wise, i.e., estimating a density map from the smallest feature map extracted. We found that for both models' small fraction of the dataset significantly contributing to its average MAE. Table 7. shows that out of the top 14 MAE contributors for W-Net, 10 of them were the top MAE contributors for CSRNet. Shanghai part A is quite biased in its partitioning. Table 8 shows the distribution of the range of number of people vs. number of images along with the results for range of number of people vs. cumulative MAE contributed. The numbers clearly show that the error rates shoot up at the biased ranges. As stated in section 1, the contemporary line of work may have eeked out an extra point to reach the top, but maybe the encoder-decoder pipeline is to be credited for this and the specialized models, including ours, may just be saturating this encoder-decoder method to its potential. The solution may just lie elsewhere.

| Count range | Sum of Absolute Errors | Percentage contribution to total loss | Total Images |
|---|---|---|---|
| 0-200 | 840.69 | 7.73% | 40 |
| 200-300 | 1911.94 | 17.57% | 48 |
| 300-500 | 2376.14 | 21.84% | 47 |
| 500-1000 | 2268.72 | 20.84% | 24 |
| >1000 | 3483.65 | 32.02% | 19 |

Table 8: Error contribution based on people count

## 8. Conclusion

In this work, we propose a U-Net inspired encoder-decoder network, W-Net, for Density map estimation and crowd counting. The inclusion of the proposed Reinforcement decoding branch helps the network converge quicker and also produce density maps with high SSIM index. With extensive experiments we report state-of-the-art results in three crowd counting datasets. Followed by our detailed ablation studies, with which we explain how and why we decided to build this network. Finally, we conclude with the error analysis stating our stand on the current solutions for the Crowd Counting problem.